\documentclass{article}
\usepackage{amssymb}

\usepackage{graphicx}
\usepackage{multirow}
\usepackage{hyperref}
\usepackage{booktabs}
\usepackage{multirow}
\usepackage{colortbl}
\usepackage{rotating}
\usepackage{amssymb}
\usepackage{amsmath}
\usepackage{graphicx}
\usepackage{enumitem}
\usepackage[preprint]{corl_2025} 
\usepackage{algorithm}
\usepackage{algpseudocode}
\usepackage[table]{xcolor}
\hypersetup{
    colorlinks=true,
    linkcolor=orange,
    filecolor=magenta,
    urlcolor=orange,
    citecolor=orange,
}

\title{CLTP: Contrastive Language-Tactile Pre-training for 3D Contact Geometry Understanding}

\author{
Wenxuan Ma$^{1, \dag}$\quad
Xiaoge Cao$^{1, \dag}$\quad 
Yixiang Zhang$^{2}$\quad 
Chaofan Zhang$^{1}$\quad
Shaobo Yang$^{3}$\quad \\
\textbf{Peng Hao}$^{4}$\quad
\textbf{Bin Fang}$^{3}$\quad 
\textbf{Yinghao Cai}$^{1}$\quad 
\textbf{Shaowei Cui}$^{1, *}$\quad 
\textbf{Shuo Wang}$^{1}$\quad \\
$^1$Institute of Automation, Chinese Academy of Sciences \quad
$^2$Beihang University \\
$^3$Beijing University of Posts and Telecommunications \quad
$^4$Samsung Research China
\vspace{-12pt}
}
\begin{document}
\maketitle
\begingroup
\renewcommand\thefootnote{} 
\footnotetext{$^\dag$Equal contribution \quad *Corresponding author: Shaowei Cui (shaowei.cui@ia.ac.cn)}
\endgroup
\vspace{-5pt}

\begin{figure}[htbp]
\centering
\includegraphics[width=\textwidth]{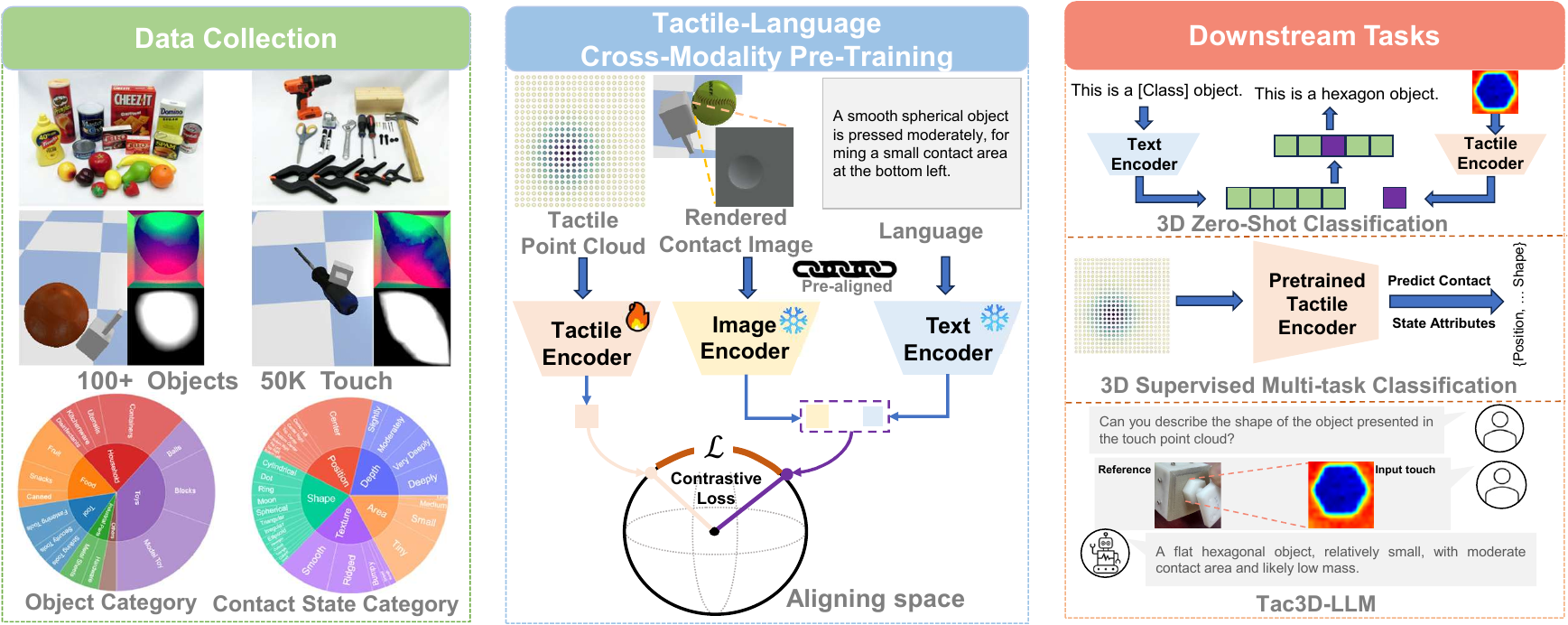}
\caption{\textbf{Overview of the CLTP framework and its downstream tasks}. The left and middle part are the TCL3D dataset construction process and the CLTP pre-training framework. CLTP uses a large multimodal model to automatically generate a detailed text description for each 2D rendered tactile image from the 3D contact deformed tactile point cloud (including contact shape, area, position, depth and texture). CLTP uses a pre-aligned and frozen vision-language feature space to achieve alignment between the three modalities (text, image and 3D tactile point cloud). After pre-training, the 3D encoder will be used for three downstream tasks.} 
\label{fig:example}
\vspace{-10pt}
\end{figure}

\begin{abstract}
Recent advancements in integrating tactile sensing with vision-language models (VLMs) have demonstrated remarkable potential for robotic multimodal perception. However, existing tactile descriptions remain limited to superficial attributes like texture, neglecting critical contact states essential for robotic manipulation. To bridge this gap, we propose CLTP, an intuitive and effective language tactile pretraining framework that aligns tactile 3D point clouds with natural language in various contact scenarios, thus enabling contact-state-aware tactile language understanding for contact-rich manipulation tasks. We first collect a novel dataset of 50k+ tactile 3D point cloud-language pairs, where descriptions explicitly capture multidimensional contact states (e.g., contact location, shape, and force) from the tactile sensor’s perspective. CLTP leverages a pre-aligned and frozen vision-language feature space to bridge holistic textual and tactile modalities. Experiments validate its superiority in three downstream tasks: zero-shot 3D classification, contact state classification, and tactile 3D large language model (LLM) interaction. To the best of our knowledge, this is the first study to align tactile and language representations from the contact state perspective for manipulation tasks, providing great potential for tactile-language-action model learning. Code and datasets are open-sourced at https://sites.google.com/view/cltp/.
\end{abstract}
\vspace{-5pt}
\keywords{Tactile-Language Representation Pre-training, Tactile Sensing, Multimodal Robot Learning} 


\section{Introduction}

    Tactile perception\cite{dahiya2009tactile,li2020review,billard2019trends}, which serves as a cornerstone of human interaction with the physical world by enabling delicate grasping\cite{romano2011human} and material recognition\cite{dai2022design}, is critical for replicating in robots the sophisticated capabilities required to master complex tasks such as precision assembly\cite{cui2021hand} and flexible object manipulation\cite{yin2021modeling,zhu2022challenges}. By interpreting real-time contact force patterns and surface deformation, robots dynamically adapt grasping postures and force application, moving beyond rigid programming\cite{sunil2023visuotactile,fang2022soft,li2024comprehensive}. This nuanced detection of contact states—including pressure distribution and geometric feedback—grants human-like adaptability, ensuring reliable physical interaction in unstructured environments\cite{cui2021toward}.

    Although significant progress has been made in the multimodal integration of vision, language, and actions in robotics\cite{radford2021learning,li2022blip,li2023blip,kim2024openvla,zhen20243d,li2025pointvla,hong2024multiply}, the cross-modal association between tactile sensing and language remains in its infancy\cite{yang2022touch,jones2025beyond}. Current research on tactile image-language alignment based on vision-based tactile sensors predominantly focuses on material tactile perception\cite{cheng2024touch100k,fu2024touch}, yet rarely provides executable linguistic descriptions directly linked to robotic manipulation scenarios – such as contact geometry information during interaction\cite{suomalainen2022survey,zhang2023gelstereo,doshi2022manipulation,liu2022novel}. In summary, tactile-language modality alignment still faces two core challenges: 1) Building robust tactile-language alignment requires sensor-agnostic representations that can generalize across hardware platforms (e.g., optical\cite{zhang2022hardware,cui2021hand,yuan2017gelsight} vs. resistive\cite{stassi2014flexible} tactile sensors) and sim2real scenarios. 2) There is currently a lack of language-annotated datasets for contact states (e.g., contact location, contact area, contact depth), which limits their practical value for robotic manipulation tasks.

    To address these challenges, we first build a dataset containing tactile 3D contact point clouds\cite{du20223d} and corresponding language labels of contact states specifically designed for robotic manipulation tasks. The dataset design embodies two core principles: First, our collected tactile sensing signals prioritize cross-platform and cross-scenario applicability. Unlike existing approaches using tactile images, we directly employ 3D point cloud data generated from physical contacts. We synthesized primary contact data through simulation\cite{wang2022tacto} while concurrently collecting real-world data using multiple visuotactile sensors\cite{yuan2017gelsight,zhang2023gelstereo}, significantly reducing acquisition costs while inherently enabling sim-to-real transfer capabilities. Second, our language annotation framework for tactile 3D point clouds deviates from conventional “tactile sensation" descriptors (e.g., "hard" or "smooth"). Instead, we implement a linguistic description through five dimensions of contact status that are critical to the manipulation task: contact location, contact area, contact geometry, penetration depth, and surface texture. This multimodal annotation system combines off-the-shelf Vision-Language Model (VLM)\cite{bai2023qwen,achiam2023gpt} description generation with manual feature engineering, resulting in comprehensive contact state characterization. The final dataset comprises 52, 425 samples covering 117 objects' interaction states with various sensors, which we called TCL3D dataset.
    
   We further propose a Contrastive Language-Tactile Pre-training (CLTP) method to align tactile 3D point clouds with natural language. Similar to other CLIP-inspired approaches for 3D point cloud alignment\cite{hegde2023clip,xue2023ulip}, CLTP leverages a frozen pre-aligned vision-language feature space to establish connections between tactile and textual modalities, achieving alignment between tactile point clouds and textual descriptions through contrastive learning. The pretraining performance of CLTP is evaluated on three downstream tasks: zero-shot 3D classification, contact state classification, and tactile 3D large language model (Tac3D-LLM) interaction. Experimental results demonstrate the necessity and effectiveness of the proposed CLTP method. We believe this contact state-aware tactile 3D point cloud encoder holds significant potential in vision-language-action (VLA) domains requiring tactile modality embedding. In summary, the primary contributions of this paper are three-fold: data, pretraining method, and LLM application.

   \begin{itemize}
       \item We present the TCL3D dataset, a large-scale paired tactile 3D point cloud-language dataset for contact deformation understanding, covering contact 3D point clouds, tactile depth images, and language descriptions in various contact scenarios, focusing on the description of contact states such as contact position, depth, and shape, etc.
       \item We propose Contrastive Language-Tactile Pre-training (CLTP) to align tactile 3D point clouds with natural language, thereby achieving tactile language understanding based on contact state understanding.
       \item Experiments demonstrate the necessity and effectiveness of TCL3D dataset and tactile 3D-language alignment in experiments at different levels, such as zero-shot tactile 3D classification and Tac3D-LLM application.
   \end{itemize}

\section{Related Work}

\subsection{Multi-modal Alignment} Agents have demonstrated immense potential in robotics\cite{jiang2022vima} and artistic creation\cite{liu2024sora} through multimodal perception and comprehension of the real world. Multimodal alignment can help machines better understand cross-modal information, enabling more efficient processing and application\cite{li2024multimodal}. Radford et al.\cite{radford2021learning} bridged vision and language modalities through self-supervised contrastive training, proposing the CLIP framework. Subsequent studies further enhanced alignment between visual and language modalities\cite{li2022blip,li2023blip,li2025benchmark,zhang2024vision}. Follow-up research has explored alignment frameworks for speech, video, depth, and other modalities with language\cite{panagopoulou2024x,zhao2024distilling}. Xue et al. developed the ULIP framework\cite{xue2023ulip,xue2024ulip}, achieving representation alignment across text, vision, and 3D point clouds. Zhu et al. successively proposed PointCLIP\cite{zhu2023pointclip,zhang2022pointclip}, integrating GPT to realize open-world 3D classification, segmentation, and detection. However, these alignment models combining point clouds, vision, and text remain focused on general 3D object structure understanding and do not align well with the contact-based local point cloud comprehension required in robotic manipulation tasks.

\subsection{Tactile-Language Pre-training} 
The alignment of tactile and language modalities has also captured the attention of roboticists\cite{yu2024octopi,hao2025tla,xu2024survey}. Fu et al.\cite{fu2024touch} designed a specialized handheld collection device equipped with DIGIT tactile sensors and webcams to compile a large-scale dataset of visual, tactile, and manually annotated/GPT-generated textual descriptions named TVL. They further proposed a tactile-visual language (TVL) model for generating textual descriptions from tactile images. Cheng et al.\cite{cheng2024touch100k} introduced Touch100K, an open dataset collecting GelSight-like tactile and visual data with corresponding text descriptions, and proposed TVL-Link to learn aligned representations of tactile, visual, and linguistic modalities. Yang et al.\cite{yang2024binding} developed UniTouch, aligning tactile modality with visual, textual, and auditory modalities. Feng et al.\cite{feng2025anytouch} further constructed the TacQuad dataset incorporating multiple sensor platforms and proposed AnyTouch to achieve cross-sensor alignment of dynamic contact sequences with visual and textual representations. However, their textual descriptions predominantly focus on material texture evaluation rather than grasping state information critical for contact-rich manipulation tasks—such as contact location, depth, and interaction dynamics. Current studies remain confined to the superficial level of tactile imagery, lacking in-depth 3D deformation understanding and cross-scenario generalization capabilities.


\section{TCL3D Dataset}

    We construct the TCL3D dataset, a comprehensive contact-state-aware tactile dataset built from YCB objects\cite{calli2015ycb}, self-made  pegs and \href{https://www.mcmaster.com/?bot=true}{McMaster}, encompassing both daily and industrial scenarios. We prepare a total of 117 objects for this dataset. According to the size of the object, we divide them into two categories: large object set (62 objects, most of which are from the YCB dataset) and small object set (55 objects). Different contact strategies were formulate to obtain the desired tactile 3D point cloud, the corresponding tactile rendering image, and the contact state-oriented tactile language description. For each contact, we create a triplet ($T_i$, $L_i$, $I_i$) of 3D tactile data, language description, and rendered contact image. Next, we introduce the data collection strategy, tactile image generation, and text description generation.
\begin{figure}[thbp]
    \centering
    \includegraphics[width=\textwidth]{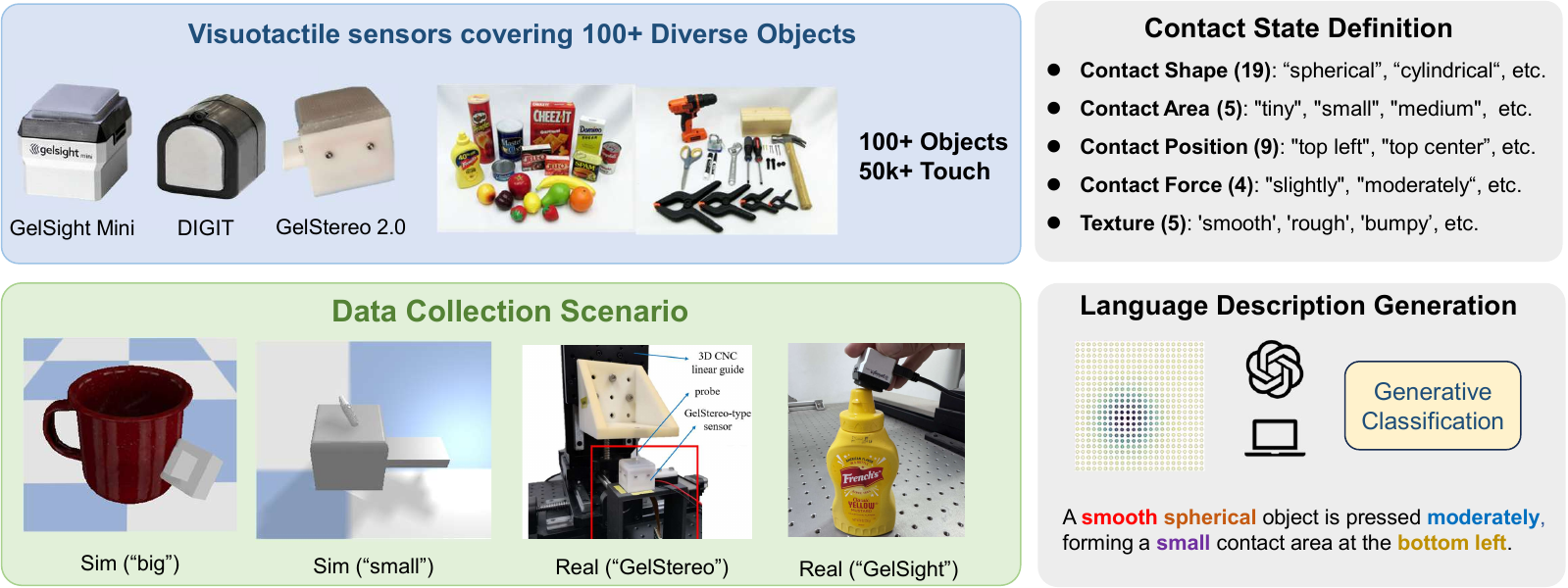}
    \caption{\textbf{TCL3D dataset:} a multimodal multi-sensor tactile alignment dataset for CLTP training and evaluation.} 
    \label{fig:dataset}
    \end{figure}
\subsection{Data Sampling}
    In order to obtain tactile point cloud data at the lowest cost, we selected the TACTO tactile simulator\cite{wang2022tacto} with strong real-time performance combined with Pybullet\cite{coumans2021pybullet} for main body simulation data acquisition, and used real GelStereo\cite{zhang2023gelstereo} and GelSight Mini\cite{yuan2017gelsight,lambeta2020digit} sensors to verify real data acquisition (Fig. \ref{fig:dataset}). In the simulated contact, we adopted different contact strategies for large objects and small objects (Fig. \ref{fig:dataset}): For large objects, we randomly sampled a point on the surface of the object, and then set a contact orientation within the 15-degree surface normal cone. The sensor moved along this orientation and applied incremental force to collect different contact depths. For small objects (less than 5 cm), we sampled an object orientation from the unit sphere, and then forced the object to contact the sensor downward and applied varying forces to generate contact samples. To obtain images that semantically align with tactile data while capturing features beyond the language modality, we synthesize images for each contact sample by converting the 3D tactile data into a mesh and rendering an RGB observation from the tactile sensor's view. 
    
    During the real data set acquisition process, we used a 3D CNC equipped with probes of different shapes to press the GelStereo sensor at different positions at different depths (Fig. \ref{fig:dataset}) to collect tactile 3D point cloud data with real contact states. At the same time, we held the GelSight sensor in contact with everyday objects and collected the corresponding tactile data to qualitatively evaluate the alignment characterization method. In addition, we use data augmentation techniques, including random sampling, point cloud translation and rotation, to expand the original collected simulated and real tactile 3D point clouds. As a result, our TCL3D dataset contains 50, 860 simulated contact samples, 1, 450 GelStereo real contact samples, and 115 GelSight contact samples.


\subsection{Natural Language Description Generation}

    We define tactile contact states comprising 19 shape categories (e.g., sphere, cuboid, triangle), 5 texture categories (e.g., smooth, ridged), 4 depth categories (e.g., slight, moderate, deep), 9 position categories (e.g., top-left, top-center, bottom-right), and 5 area categories (e.g., tiny, small, medium, large) (Fig. \ref{fig:dataset}). Object shape and surface texture are characterized using GPT-4o\cite{hurst2024gpt}, while contact position, depth, and area are directly calculated from the tactile point cloud data. On average, 400–500 contact samples per object are collected to capture a range of diverse contact scenarios.
    
    We leverage the metadata acquired from contact sampling to generate text descriptions. The metadata includes a series of words, each identifying a contact state of the contact (e.g., shape, position, force). We adopt a simple prompt to construct meaningful sentences that are used during pretraining, such as: 'a [Texture] [Shape] object, pressed [Depth] in [Position] with [Area] contact area.'  We discretize each dimension (e.g., depth) into specific words. The generated text is then input into a pretrained text encoder to obtain a contact-state-aware language representation.
\begin{figure}[thbp]
    \centering
    \includegraphics[width=\textwidth]{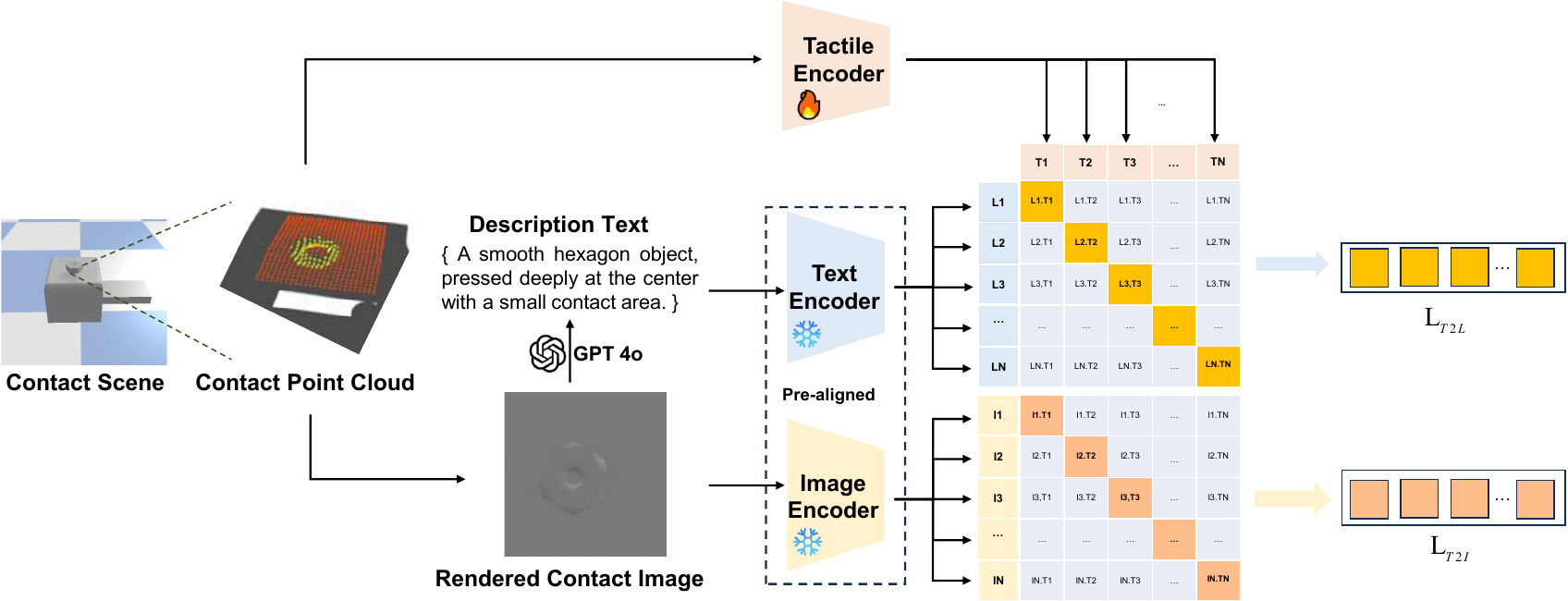}
    \caption{\textbf{Architecture of CLTP framework.} It leverages a frozen pre-aligned vision-language feature space to establish connections between tactile and language modalities.}
    \label{fig:method}
    \end{figure}

\section{Method}
    We aim to learn a unified tactile representation for contact-rich manipulation tasks that captures critical contact states across touch and other modalities, such as language and vision. First, we conduct language-tactile pretraining, utilizing 3D tactile input and our novel language descriptions, which explicitly capture multidimensional contact states. Subsequently, we show how our learned representation can be applied to various tactile downstream tasks.


    CLTP aligns the 3D tactile data, 2D depth images and comprehensive contact states to a unified feature space. We achieve that by aligning 3D touch emebddings to a pre-aligned image-language feature space using contrastive learning as shown in Fig. 
    \ref{fig:method}.

    During pre-training, given a 3D tactile sample, we extract the language feature $f^L=E_L(L)$ and image feature $f^I=E_I(I)$ based on the pre-aligned and frozen lanuage encoder and image encoder in CLIP\cite{radford2021learning}. We aim to train 3D tactile encoder $E_T$ such that its 3D feature $f^T=E_T(T)$ is aligned with its language and image features. We formulate the Tactile-Language alignment using the CLIP-Style contrastive loss:

   \begin{equation}
   \mathcal{L}_{\mathrm{T2L}}=-\frac{1}{2}\sum_i\log\frac{\exp(\mathbf{f}_i^\mathbf{T}\mathbf{f}_i^\mathbf{L}/\tau)}{\sum_j\exp(\mathbf{f}_i^\mathbf{T}\mathbf{f}_j^\mathbf{L}/\tau)}+\log\frac{\exp(\mathbf{f}_i^\mathbf{T}\mathbf{f}_i^\mathbf{L}/\tau)}{\sum_j\exp(\mathbf{f}_j^\mathbf{T}\mathbf{f}_i^\mathbf{L}/\tau)}
   \end{equation}
    where i, j are the sampling indices, and $\tau$ is a learnable temperature parameter. 
    However, text is not able to describe fine-grained shape and texture features, which are critical for enhancing our 3D tactile representation. So we introduce Tactile-Image alignment loss as: 
     \begin{equation}
     \mathcal{L}_{\mathrm{T2I}}=-\frac{1}{2}\sum_i\log\frac{\exp(\mathbf{f}_i^\mathbf{T}\mathbf{f}_i^\mathbf{I}/\tau)}{\sum_j\exp(\mathbf{f}_i^\mathbf{T}\mathbf{f}_j^\mathbf{I}/\tau)}+\log\frac{\exp(\mathbf{f}_i^\mathbf{T}\mathbf{f}_i^\mathbf{I}/\tau)}{\sum_j\exp(\mathbf{f}_j^\mathbf{T}\mathbf{f}_i^\mathbf{I}/\tau)}  \end{equation}
    Our final training objective is to train the 3D tactile encoder $E_T$ that minimizes the sum of two contrastive alignment losses above:
    
    \begin{equation}    \min_{E_{\mathrm{T}}}\mathcal{L}_{\mathrm{T2L}}+\mathcal{L}_{\mathrm{T2I}}.  
    \end{equation}
    
    
   
    \section{Experiments}

    We conduct a series of experiments to answer the following key research questions:
    \begin{itemize}
    \item Compared with the existing point cloud-language alignment framework, does our proposed CLTP have significant advantages in 3D deformation understanding?
    \item CLTP uses a specific tactile 3D point cloud modality. How is its sim2real generalization performance?
    \item After aligning the tactile and language modalities for contact deformation attributes, what new possibilities are there in combining with a large language model?
\end{itemize}
    To answer the above questions, we evaluate the performance of our representation alignment based on CLTP in combination with three downstream tasks, including zero-shot touch classification, standard contact state classification, and Tac3D-LLM.

\textbf{Implementations.} We base out model on ULIP-2\cite{xue2024ulip}. For the 3D tactile input, we uniformly sample $N_p = 1024$ points for both training and testing. We train and evaluate our model on the TCL3D dataset. For different tactile sensors, we normalize their scales to a unified spatial range. To evaluate the generalization performance of our modal, we also evaluate it with real-word data collected by two sensors: GelSight Mini\cite{{yuan2017gelsight,lambeta2020digit}} and GelStereo\cite{zhang2023gelstereo}. 

\subsection{Zero-shot 3D Touch Classification}
We evaluate CLTP with zero-shot classification tasks, enabled by the emergent alignment with language during pretraining, to demonstrate how our tactile features align with language descriptions. Note that our contact states consist of several dimensions, which lead to a large number of possible combinations. To better illustrate how our features align at each description dimension, we conduct 3D zero-shot classification on each dimension. We compute the cosine similarity between tactile embeddings and their corresponding text prompts. Class predictions are chosen based on the highest scores, without training on labeled data. We conduct zero-shot tactile classification by prompting the model with "This is a [Shape]," where [Shape] is the shape description, and "Contact is [Depth]," "Contact at [Position]," etc., where [Depth] and [Position] are respectively the contact depth and position. Table \ref{tab:zero_shot_cls} shows the results of our method. Our zero-shot method shows comparable performance, which not only indicates a strong tactile representation that is well-aligned with text but also demonstrates that off-the-shelf models trained for other modalities can be successfully used to solve touch-sensing tasks.

\begin{table}[]
\centering
\begin{tabular}{cccccc}
\hline
Methods        & Shape         & Texture       & Depth         & Position      & Area          \\ \hline
CLTP(w/o image) & 52.6          & 65.7          & 96.5          & 91.9          & 67.1          \\
\rowcolor[HTML]{DAEFF2}CLTP           & \textbf{70.1} & \textbf{68.9} & \textbf{98.3} & \textbf{94.4} & \textbf{81.8} \\ \hline
\end{tabular}
\caption{\textbf{Tactile contact state zero-shot classification(Metric: accuracy\%). } } 
\label{tab:zero_shot_cls}
\end{table}

\begin{figure}[!t]
    \centering
    \includegraphics[width=0.8\textwidth]{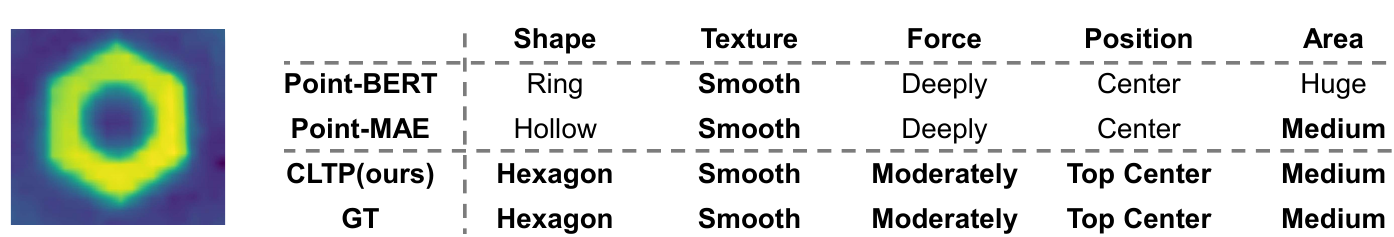}
    \caption{\textbf{Case Study:} CLTP is more sensitive to the shape, pressing depth, and contact position of the contact 3D point cloud than the baseline models.}
    \label{fig:case1}
    \end{figure}
\subsection{Standard Contact State Classification}
We present 3D classification results on our TCL3D dataset and real-world tactile data collected from GelStereo\cite{zhang2023gelstereo} and GelSight Mini\cite{yuan2017gelsight,lambeta2020digit} sensors. We pretrain all backbone models on the TCL3D dataset, then freeze the encoder and finetune MLPs to conduct classification for each tactile attribute, including shape, position, force, etc. 

As shown in Table \ref{table:contactstatecls}, our proposed CLTP achieves the best performance across all tactile attributes on both the synthetic and real-world datasets.  especially in shape and texture,  CLTP achieves over 95\% success in contact force, texture, position, and area, and an 84.8\% shape classification accuracy on the TCL3D dataset, compared to 61.2\% by the CLTP variant without image modality, and significantly outperforms Point-BERT\cite{yu2022point} (28.7\%) and Point-MAE\cite{pang2022masked} (31.6\%). On real-world data, CLTP also achieves a 71.2\% in shape recognition, surpassing Point-MAE by 47.3\%, and the CLTP variant by 15.9\%, exhibiting strong transferability in real-world scenarios. We attribute these improvements to language-based supervision helps the model understand high-level contact semantics, the visual features from contact images provide richer shape and texture cues. By concurrently aligning with language and visual modalities, CLTP learns a tactile feature space that effectively captures critical contact states while enhancing the characterization of surface shape and texture. We also conduct comparative experiments on vision-based tactile pre-training methods, see Appendix \ref{secA}.

\begin{figure}[!t]
    \centering
    \includegraphics[width=1.0\textwidth]{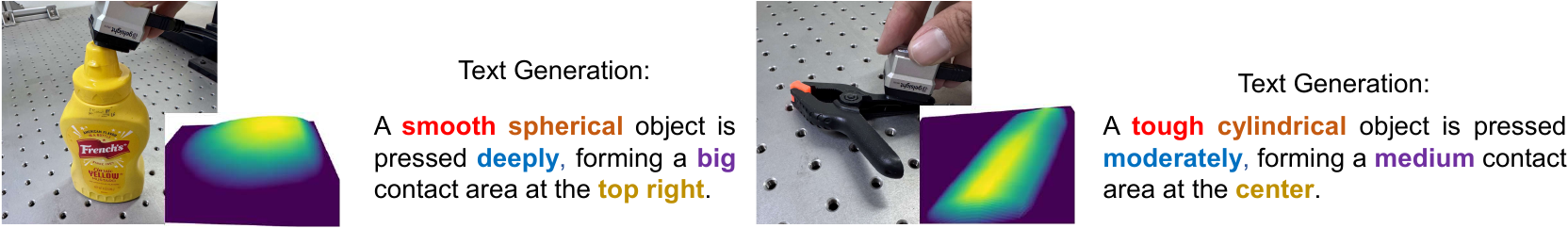}
    \caption{\textbf{A case study of generating text descriptions from realistic tactile point clouds.}}
    \label{fig:case2}
    \end{figure}

\begin{figure}[!t]
    \centering
    \includegraphics[width=0.8\textwidth]{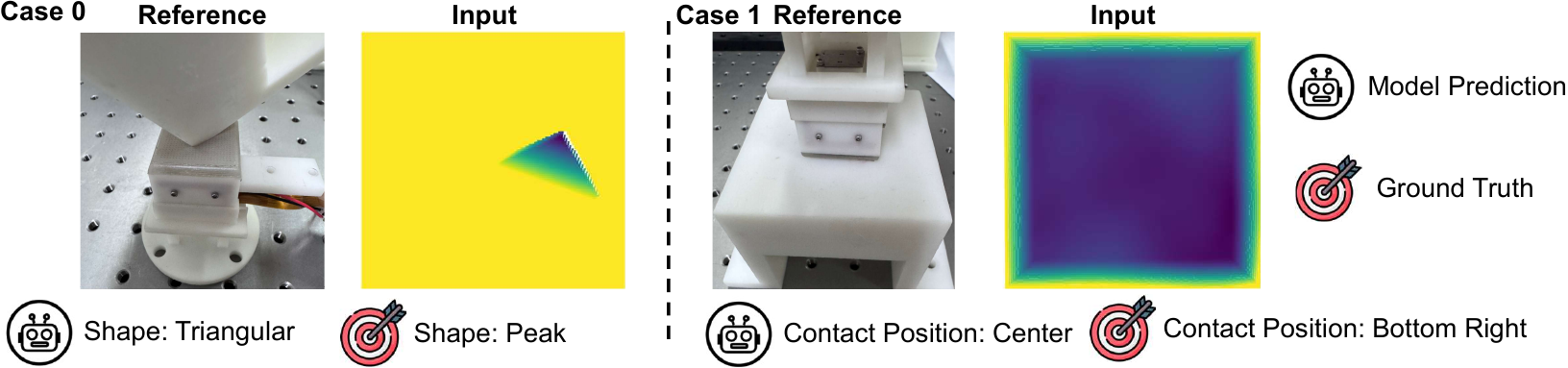}
    \caption{\textbf{Analysis of failed cases of CLTP text description generation for tactile point clouds.} Case 1: shape description ambiguity; Case 2: large plane contact leads to description noise.}
    \label{fig:case3}
    \end{figure}


We analyze failure cases in our experiments and find several common reasons, such as the shape’s ambiguity, because our 19 shape classes may exhibit some overlap, such as triangle and peak, sphere and ellipsoid, etc. The model may encounter semantic ambiguity when processing these shapes. Meanwhile, we identify the deepest contact point as the contact position, this may lead to diverse labels when facing large flat objects, as shown in Fig. \ref{fig:case3}. Overall, CLTP exhibits strong performance across multiple tactile understanding dimensions and demonstrates robust generalization to real-world scenarios, providing great potential for contact-rich manipulation tasks.


\begin{table}[!t]
\resizebox{\textwidth}{!}{
\begin{tabular}{ccccccccccc}
\hline
\multirow{2}{*}{Methods} & \multicolumn{5}{c}{TCL3D Dataset}                                             & \multicolumn{5}{c|}{Real World Data}                                          \\ \cline{2-11} 
                         & Shape         & Texture       & Depth         & Position      & Area          & Shape         & Texture       & Depth         & Position      & Area          \\ \cline{2-11} 
Point-BERT\cite{yu2022point}               & 28.7          & 89.5          & 64.4          & 80.8          & 91.3          & 21.9          & 91.1          & 52.1          & 51.2          & 73.4          \\
Point-MAE{\cite{pang2022masked}}                & 31.6          & 90.2          & 68.1          & 82.9          & 92.6          & 23.9          & 89.2          & 52.1          & 55.3          & 56.2          \\
CLTP (Ours w/o image)          & 61.2          & 94.5          & 94.7          & 91.4          & 94.8          & 55.3          & 90.3          & 77.3          & 73.8          & 64.4          \\
\rowcolor[HTML]{DAEFF2}CLTP (Ours)                     & \textbf{84.8} & \textbf{96.1} & \textbf{99.2} & \textbf{95.9} & \textbf{97.1} & \textbf{71.2} & \textbf{92.7} & \textbf{83.6} & \textbf{81.4} & \textbf{77.2} \\ \hline
\end{tabular}}
\caption{\textbf{Standard tactile contact state classification}.  We compare our 3D tactile features with other methods. The metric is accuracy(\%). }
\label{table:contactstatecls}
\end{table}

\subsection{Tac3D-LLM}\label{sec:tac3dllm}
\begin{figure}[htbp]
\centering
\includegraphics[width=0.8\textwidth]{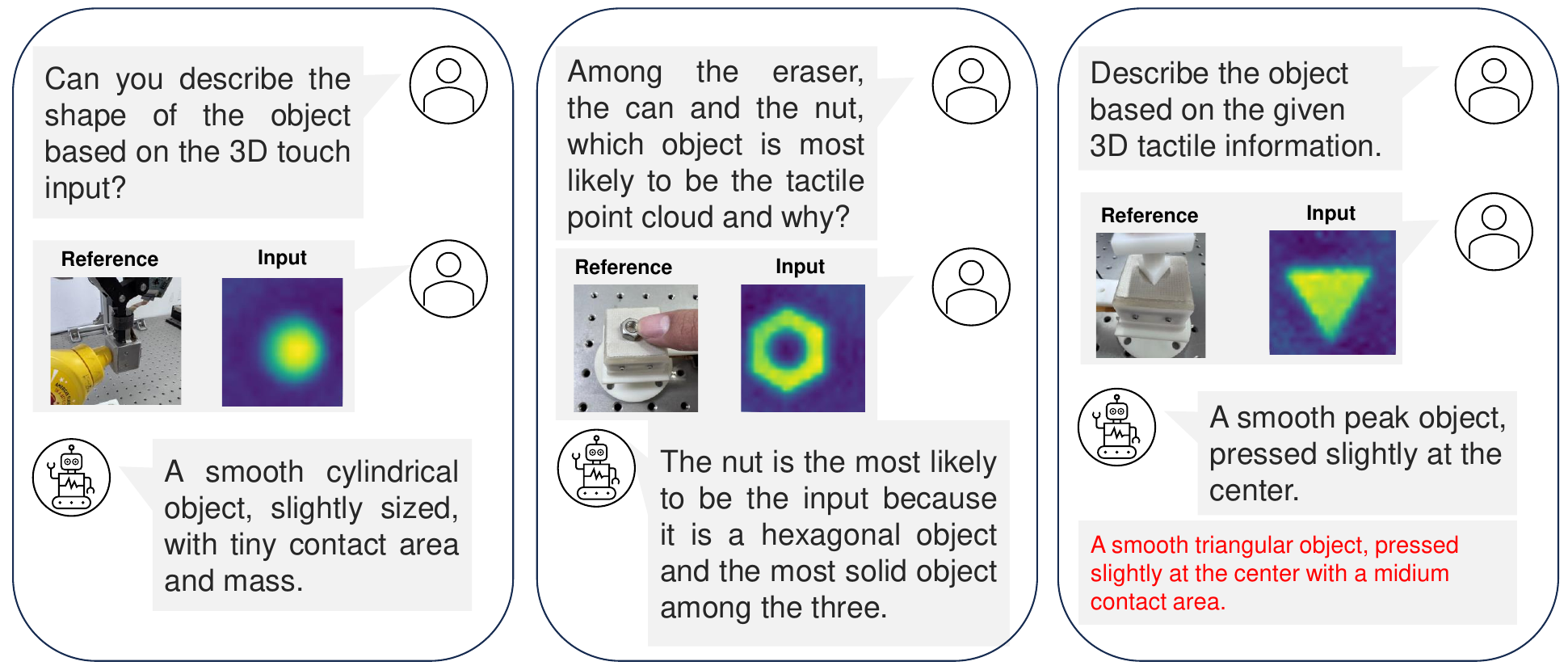}
\caption{\textbf{Tac3D-LLM.} Our Tac3D-LLM is capable of performing a range of tactile question-answering tasks, such as contact description and multi-modal reasoning. }
\label{fig:tac3d-llm}
\end{figure}

By leveraging a semantically aligned feature space and a large language model (LLM), the system can perform multi-modality reasoning. We utilize Qwen2.5-VL-3B \cite{bai2023qwen} as our multimodal large language model (MLLM) backbone and finetune a simple MLP network to align tactile features with LLM tokens through a captioning task, following LLaVA \cite{liu2023visual}. With an aligned feature space, our Tac3D-LLM can learn to understand the tactile modality and perform reasonable inference, leveraging its knowledge to reason and provide accurate responses, demonstrates the potential of integrating large language models (LLMs) with contact-rich manipulation tasks. We show some example tasks in Fig. \ref{fig:tac3d-llm}. We further demonstrate the application of Tac3D-LLM in the strawberry precise positioning and delicate grasping experiment, see Appendix \ref{secB} for additional details.

We compare our model with two baselines with different tactile encoders: PointMAE-LLM and PointBERT-LLM. We find that the baseline models can only answer the specific types of questions they were trained on and perform with limited generalization. In contrast, Tac3D-LLM generalizes effectively to unseen questions. We hypothesize that this is because, without alignment, the correspondence from such an encoder to the output of the LLM may only be a monomorphism—that is, it maps inputs to outputs in a one-to-one manner, but not necessarily in a meaningful or generalizable way. As a result, the model can only generate responses for the specific problem it was trained on. Furthermore, the unaligned encoder incurs significantly higher loss compared to the aligned tactile encoder, highlighting the critical importance of feature alignment in multimodal learning. Quantitative results are provided in Appendix \ref{secC}.


\section{Conclusion}
\label{sec:conclusion}
In this study, we introduce CLTP, a tactile 3D point cloud and language alignment representation model for understanding 3D contact deformation in robotic manipulation tasks. We first construct the TCL3D dataset, and collect 50k+ point cloud-tactile image-text samples by collecting various configurations of 117 objects with 3 different sensors in simulation and reality. Each sample contains a text description of the contact point cloud in five dimensions: contact shape, area, depth, position, and texture. With the help of the existing image-text pre-training model, we use contrastive learning to align the tactile 3D point cloud modality with the text description. The experimental results show that compared with direct representation alignment learning, its image modality bridging method has achieved a significant improvement in the generation of contact states in all five dimensions (especially contact shape, 52.6\% vs 70.1\%).

Experimental results show that compared with existing point cloud-text alignment models, CLTP has a dominant advantage in 3D contact deformation understanding (84.8\% vs 28.7\%/31.6\% in terms of contact shape). Additionally, we further verify the sim2real capability of CLTP, and the results show that CLTP Encoder, which is mainly trained with simulated point cloud data, can generalize well to daily contact data collected by real sensors. We also try to connect the trained tactile 3D point cloud encoder to LLM for 3D contact deformation language understanding, and demonstrated the possibility of combining CLTP with LLMs reasoning in daily contact samples and a strawberry tactile positioning and delicate grasping task.

\clearpage
\section{Limitations}
\label{sec:limitations} 
Despite some initial results, CLTP still has some areas that need to be improved and explored. First, due to the cost of data collection, we chose the TACTO simulator with high real-time performance but average accuracy. In order to simulate a more realistic tactile 3D deformation, a higher-precision tactile simulator is worth considering, such as Taxim\cite{si2022taxim}, Tacchi\cite{chen2023tacchi}, etc. Secondly, CLTP currently lacks accurate modeling of numbers, which makes the contact state of each dimension at the qualitative evaluation level. It is more important to quantitatively characterize the language description of each dimension such as contact position, contact area, contact depth, and align the representations for the operation state. Finally, at the point cloud level, the fusion of visual and tactile modalities and the alignment with the language modality are crucial for robot multimodal perception and VLA series generalized manipulation skill learning. We welcome the community to follow our progress and work together to address the above challenges.	

%

\bibliography{example}  

\clearpage
\appendix
\renewcommand{\thetable}{\arabic{table}}
\section*{Appendix}
\setcounter{page}{1}

This supplementary material provides additional details on the proposed method and experimental results that could not be included in the main manuscript due to page limitations.
Specifically, this appendix is organized as follows.
\begin{itemize}[left=1em]
    \item Sec.~\ref{secA} presents additional comparative results of  vision-based tactile pre-training methods. 
    \item Sec.~\ref{secB} presents detailed real-world applications of Tac3D-LLM on grasping task. 
    \item Sec.~\ref{secC} presents quantitative performance comparison for Tac3D-LLM and baselines.
\end{itemize}

\section{Comparative Results of Vision-based Tactile Pre-training Methods}
\label{secA}

We evaluate our model's performance by comparing it with two state-of-the-art vision-based tactile pre-training methods, Unitouch\cite{yang2024binding} and TVL\cite{fu2024touch}. Using pre-trained and frozen tactile features, we employ multi-layer perceptrons (MLPs) for standard contact state classification, with results presented in Tab. \ref{tab:rgb_method}. Vision-based tactile models, which focus on surface texture and material classification, demonstrate good performance in tasks dominated by visual cues, achieving high accuracies in Texture (94.7\% and 95.3\%) and Area (92.3\% and 94.4\%). However, when addressing contact-state attributes, such as contact force and position, the performance of vision-based tactile models declines significantly, with accuracies of 59.3\% and 65.6\% in Depth, 65.8\% and 63.4\% in Shape, and 85.2\% and 88.9\% in Position. In contrast, our contact-state-aware approach, which leverages comprehensive 3D tactile data, achieves remarkable improvements across all tasks, attaining accuracies of 99.2\% in Depth, 84.8\% in Shape, and 95.9\% in Position, demonstrating superior generalization capability across a variety of visuo-tactile sensing platforms.

\begin{table}[htbp]
\centering
\begin{tabular}{cccccc}
\hline
Methods             & Shape         & Texture       & Depth         & Position      & Area          \\ \hline
Unitouch            & 65.8          & 94.7          & 59.3          & 85.2          & 92.3          \\
TVL                 & 63.4          & 95.3          & 65.6          & 88.9          & 94.4          \\

\rowcolor[HTML]{DAEFF2}CLTP(ours) & \textbf{84.8} & \textbf{96.1} & \textbf{99.2} & \textbf{95.9} & \textbf{97.1} \\ \hline
\end{tabular}
\caption{\textbf{Standard tactile contact state classification}.  We compare our 3D tactile features with two tactile vision-based tactile models. We frozen the models and train multiple MLPS to perform classification. The metric is accuracy(\%). }
\label{tab:rgb_method}
\end{table}

\section{Tac3D-LLM Grasping Refinement}
\label{secB}

In Section \ref{sec:tac3dllm}, we integrated Contrastive Language-Tactile Pretraining (CLTP) with a Large Language Model (LLM) to enable tactile sensing and reasoning, which we called Tac3D-LLM. We further evaluate its effectiveness through experiments on precise strawberry positioning and delicate grasping. This task tackles two critical challenges: executing precise positioning to prevent the strawberry from slipping and applying accurate force to avoid damaging the strawberry. We employ a 3D contact point cloud as input and leverage our Tac3D-LLM framework to enable reasoning and adjustment. Several predefined action primitives, including "move downward" and "increase force," are utilized for robot control. The closed-loop control procedure is illustrated in Algorithm~\ref{alg:llm_st}.

\begin{figure}[thbp]
    \centering
    \includegraphics[width=\textwidth]{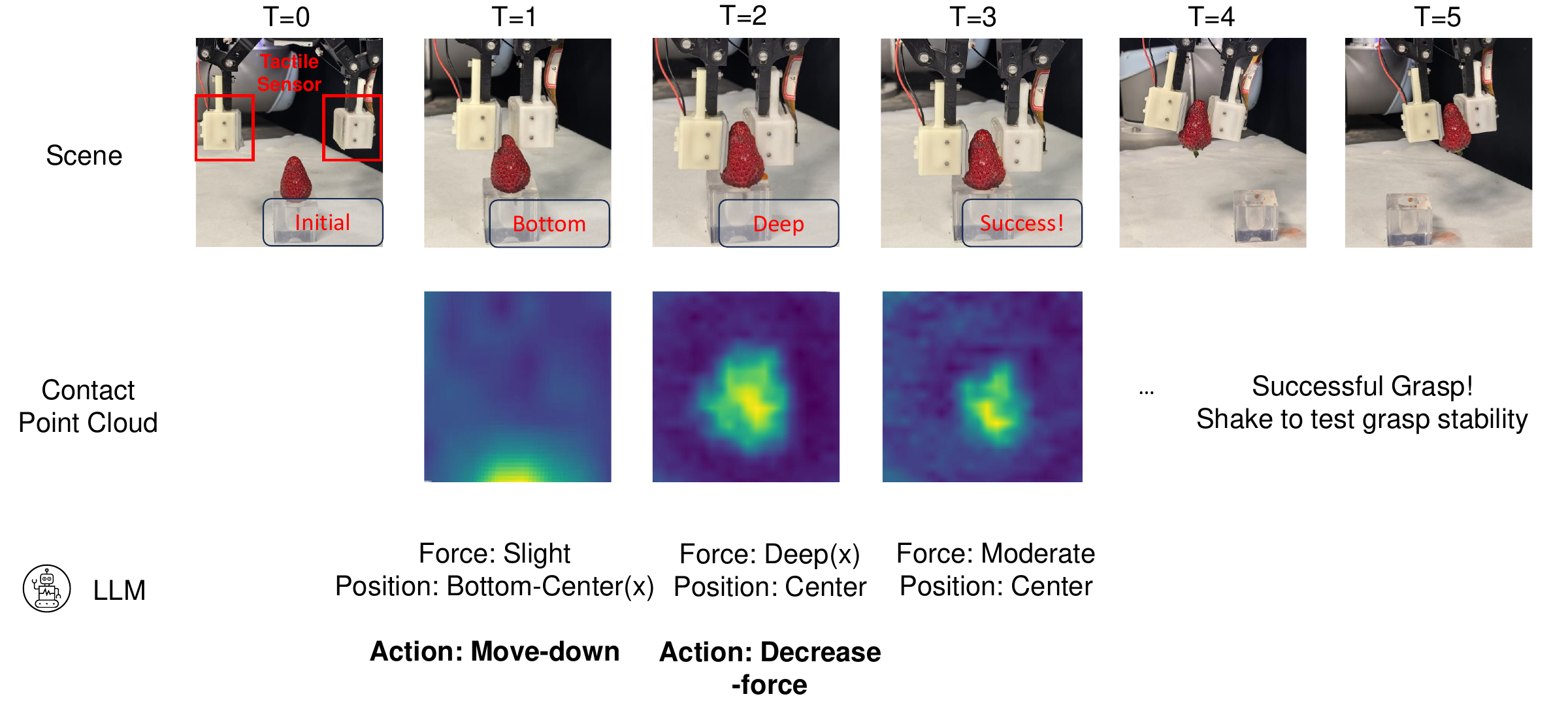}
    \caption{\textbf{Tac3D-LLM Application.} Firstly, the robotic arm predicts and executes a grasp. Tac3D-LLM then processes 3D tactile data from sensors, along with task instructions and goals, to assess the contact state and determine appropriate actions. For instance, at T=1, the sensor detects a bottom-center contact position, suggesting potential grasp instability, prompting the arm to move downward. At T=2, the sensor detects a centered contact but excessive force, leading to a signal to reduce the applied force. Through two iterative closed-loop adjustments, the gripper achieves a stable grasp.}
    \label{fig:llm_st}
    \end{figure}

Tac3D-LLM enables stable grasping of a strawberry by processing real-time 3D contact states and using compositional reasoning to infer spatially grounded adjustments, such as "move the gripper downward" or "slightly reduce gripping force" (see Fig. \ref{fig:llm_st}). The robot executes these adjustments using predefined action primitives, iterating in a closed-loop process until the contact state aligns with the target goal. Tac3D-LLM excels in contact-state-aware manipulation by autonomously optimizing physical interactions through iterative tactile signal feedback and LLM planning. This is driven by: (1) \textbf{State-aware grounding}, which maps tactile observations to semantically interpretable contact features, and (2) \textbf{Goal-conditioned reasoning}, which leverages the LLM’s compositional inference to align current contact states with desired objectives.

\begin{algorithm}[h!]
\caption{Tac3D-LLM-Guided Tactile Grasp Refinement}\label{alg:llm_st}
\label{alg:tactile_grasp}

\begin{algorithmic}[1]

\State \textbf{Inputs:} Tactile point cloud \( T \), natural language instruction \( L \), Tac3D-LLM model \( M \), action set \( A \)
\State \textbf{Output:} End-effector action

\State \textbf{Initialize:} Robot arm and tactile-enabled gripper
\State \textbf{Define} action primitives:
\[
A = \{\texttt{move\_up}, \texttt{move\_down}, \texttt{move\_left}, \texttt{move\_right},
\texttt{increase\_force}, \texttt{decrease\_force}, \texttt{reset}\}
\]
\State \textbf{Define} grasp success descriptors:
\[
K = \{\text{"Stable"}, \text{"Appropriate"}, \text{"Secure"}, \text{"Reliable"}\}
\]

\State \textbf{Set} \texttt{Success} \( \gets \textbf{False} \)

\While{\textbf{not} \texttt{Success}}
    \State \textbf{Input} current tactile signal and task prompt into Tac3D-LLM:
    \[
    D_t \gets M(T_t, L)
    \]
    \State \textbf{Extract} action from the large language model (LLM):
    \[
    a_t \gets \text{Extract}(D_t, A)
    \]
    \State \textbf{Execute} action \( a_t \) to control the end effector

    \If{\texttt{any descriptor in } \( K \) \texttt{ appears in } \( D_t \)}
        \State \texttt{Success} \( \gets \textbf{True} \)
        \State \Return \textbf{"Grasp completed successfully"}
    \EndIf

\EndWhile
\end{algorithmic}
\end{algorithm}

\section{Quantitative Performance Comparison for Tac3D-LLM}
\label{secC}
We quantitatively compare our Tac3D-LLM model against two baselines, PointMAE-LLM and PointBERT-LLM, on the 3D tactile captioning task, using identical touch point cloud and text prompts. We evaluate the models using 1000 randomly sampled 3D tactile data points from the TVL3D dataset as ground truth. Following \cite{yang2024binding}, we employ GPT-4 for automatic evaluation, instructing it to score each model's outputs on a 1–5 scale based on the reference response. As shown in Tab. \ref{tab:llm_quant}, Tac3D-LLM significantly outperforms the baseline LLMs, demonstrating superior understanding of 3D tactile signal.

\begin{table}[tbp]
\centering
\begin{tabular}{ccc}
\hline
\multirow{2}{*}{Method}  & \multirow{2}{*}{LLM} & Eval             \\ \cline{3-3} 
                         &                      & GPT-4 Rating (↑) \\ \hline
PointBERT-LLM                & Qwen2.5-VL           & 1.73             \\
PointMAE-LLM                 & Qwen2.5-VL           & 1.61             \\
\rowcolor[HTML]{DAEFF2}Tac3D-LLM(ours) & \textbf{Qwen2.5-VL}  & \textbf{4.28}    \\ \hline
\end{tabular}
\caption{\textbf{Tactile contact states caption evaluation}.  We evaluate our Tac3D-LLM and two baselines on our test cases from TVL3D. Each model's output is rated by GPT-4 from 1 to 5. }
\label{tab:llm_quant}
\end{table}

\end{document}